\documentclass[]{spie}  
\usepackage[]{graphicx}
\usepackage[]{hyperref}

\usepackage{amsfonts}

\title{Deep learning and face recognition: the state of the art.}

\author{
Stephen Balaban \\
Lambda Labs \\
\href{mailto:s@lambdalabs.com}{s@lambdalabs.com} \\
May 15th, 2015
}

\authorinfo{Author Information: s@lambdalabs.com, www.lambdalabs.com}

\begin{document} 
\maketitle 

\begin{abstract}

Deep Neural Networks (DNNs) have established themselves as a dominant technique
in machine learning. DNNs have been top performers on a wide variety of tasks
including image classification, speech recognition, and face recognition.
\cite{ILSVRCarxiv14, hinton2012deep, sun2014deep}  Convolutional neural
networks (CNNs) have been used in nearly all of the top performing methods on
the Labeled Faces in the Wild (LFW) dataset. \cite{lfw, schroff2015facenet,
taigman2013, sun2014deep} In this talk and accompanying paper, I attempt to
provide a review and summary of the deep learning techniques used in the
state-of-the-art. In addition, I highlight the need for both larger and more
challenging public datasets to benchmark these systems.

Despite the ability of DNNs and autoencoders to perform unsupervised feature
learning, modern facial recognition pipelines still require domain specific
engineering in the form of re-alignment.  For example, in Facebook's recent
DeepFace paper, a 3D ``frontalization" step lies at the beginning of the
pipeline. This step creates a 3D face model for the incoming image and then
uses a series of affine transformations of the fiducial points to ``frontalize"
the image. This step enables the DeepFace system to use a neural network
architecture with locally connected layers without weight sharing as opposed to
standard convolutional layers. \cite{taigman2013} Deep learning techniques
combined with large datasets have allowed research groups to surpass human
level performance on the LFW dataset. \cite{sun2014deep, schroff2015facenet}

The high accuracy (99.63\% for FaceNet at the time of publishing) and
utilization of outside data (hundreds of millions of images in the case of
Google's FaceNet) suggest that current face verification benchmarks such as LFW
may not be challenging enough, nor provide enough data, for current
techniques.\cite{schroff2015facenet, sun2014deep} There exist a variety of
organizations with mobile photo sharing applications that would be capable of
releasing a very large scale and highly diverse dataset of facial images
captured on mobile devices. Such an ``ImageNet for Face Recognition" would
likely receive a warm welcome from researchers and practitioners alike.

\end{abstract}



\section{INTRODUCTION}

The application of deep learning and representation learning to the domain of
facial recognition has been the driving force behind recent advances in the
state-of-the-art. The most accurate techniques of today leverage datasets of
increasing size in conjunction with convolutional neural networks.\cite{lfw,
ILSVRCarxiv14}


This paper presents a brief historical overview of face recognition, an
overview of the field of representation learning and deep learning, a look at
how those fields are influencing the state-of-the-art of face recognition, and
proposes future work to develop a new benchmark and dataset for face
recognition research.

\section{BACKGROUND}

Previous image recognition and facial recognition pipelines relied on
hand-engineered features such as SIFT, LBP, and Fisher vectors. One only needs
to look at the techniques used in the ImageNet Large Scale Visual Recognition
Challenge (ILSVRC) in 2011 to see engineered features at the top of the list.
\cite{ILSVRCarxiv14}

This changed at the end of 2011, first with a demonstration by Le et al.~of
large scale feature learning with a sparse autoencoder. This autoencoder was
trained using asynchronous Stochastic Gradient Descent (SGD) on 1,000 machines
(16,000 cores) at Google using data from YouTube. It was then used to
initialize the weights of a DNN which set a new record for performance on
ImageNet.\cite{le2012} Then, in 2012, Krizhevsky, Sutskever, and Hinton showed that networks
with similar performance could be trained with one computer and two GPUs. Their
``SuperVision" network resulted in a 37\% gain over competing hand engineered
features in the 2012 ILSVRC and paved the way for large scale feature learning
without large scale server infrastructure.\cite{alexnet}  This marked the beginning of deep
learning as the de-facto feature extraction algorithm in the field of visual
object recognition. These advances were applied with much success to facial
recognition by Taigman, et al.~in their 2014 ``DeepFace" paper.\cite{taigman2013}

\section{DEEP LEARNING}

\subsection{Feature Engineering vs. Feature Learning}

Computer vision and signal processing algorithms often have two steps: feature
extraction, followed by classification. For example, an image feature
extractor, such as Haar, SIFT, or SURF, takes in raw input in the form of
pixels and transforms it into a feature vector that can be classified by a
general classification algorithm such as a Support Vector Machine (SVM).  These
feature extractors are custom-built, ``engineered" features. In LBP facial
recognition, weights are applied to specific portions of the face to emphasize
or de-emphasize certain regions.\cite{ahonen2006} This type of feature
engineering results in brittle, over-specialized feature extractors that cannot be
applied to other problem domains, let alone other modalities. Feature
learning methods, on the other hand, \textit{learn} a feature extractor based
on the statistics of the training data and have been successfully applied to a
variety of different domains and modalities. \cite{ILSVRCarxiv14, hinton2012deep, sun2014deep}

\begin{figure}[h]
   \begin{center}
   \begin{tabular}{c}
   \includegraphics[height=215px]{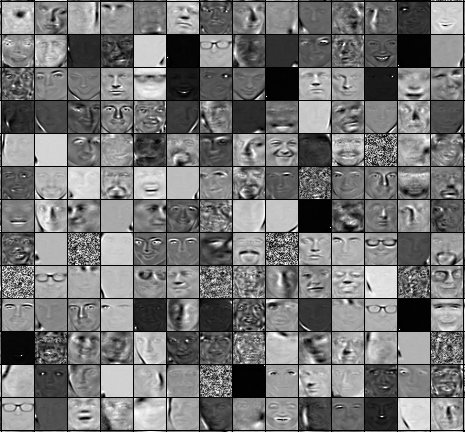}
   \end{tabular}
   \end{center}
   \caption[Autoencoder Features]{Features learned by training a Stacked
Denoising Autoencoder with unlabeled face data. Note the rich variety of
learned features including glasses, facial hair, toothy smile, and even
sunglasses.}
\end{figure}

\subsection{A quick introduction and review of gradient based learning}

Deep learning and feature learning problems share a common structure:

\begin{enumerate}
\item The creation of some layered architectures parameterized on $\theta$. (Autoencoders, DNNs, CNNs, etc.)
\item The definition of a loss functional\footnote{The loss functional is a functional because it is a function from a vector space onto its underlying scalar field.}, $\mathcal{L}$.
\item The minimization of the loss functional with respect to $\theta$ using an optimization algorithm and training data.
\end{enumerate}

Gradient based optimization methods are commonly used to perform the
optimization in step 3. \cite{bengio2012practical, bengio2013representation}
Stochastic Gradient Descent (SGD) is a modification of ``batch" gradient
descent where parameter updates are made after calculating a stochastic
approximation of the gradient. This approximation is made using a random subset of the training data,
$\mathbf{m} \subseteq x$ of size $B$, called a \textit{mini-batch}.
The assumption made by SGD is that the stochastic approximation of the
gradient, which is a random variable, has the same \textit{expected} value as
the deterministic gradient. $B$ is also known as the \textit{batch size}.  The
parameters being optimized, $\theta$, are updated after each mini-batch using
the formula below.

$$
\theta_{t+1} \leftarrow \theta_{t} - \epsilon(t) \frac{1}{B} \sum_{b=0}^{B - 1}\frac{\partial \mathcal{L}(\mathbf{m}_b, \theta)}{\partial \theta}
$$

The reason that we average the gradients as opposed to simply summing is due to
the fact that the sum of the mini-batch gradients would change significantly as
a function of $B$, thus changing the optimal learning rate schedule $\epsilon$.
By averaging, the variance of the stochastic approximation of the gradient
becomes inversely proportion to $B$. So, if $B$ decreases, the variance of the
approximation goes up which would lead to a slight reduction of the optimal
learning rate, and vice versa for an increase in $B$.\cite{bengio2012practical}
Different values of $B$ lead to different types of gradient descent:

\begin{enumerate}
\item Online stochastic gradient descent: $B = 1$.
\item Mini-batch stochastic gradient descent: $B > 1$ but $B < |x|$.
\item ``Batch" gradient descent, $B = |x|$. Note that in this case the gradient is no longer a random variable and is deterministic.
\end{enumerate}

For a more thorough treatment of the topic of training deep architectures with
gradient descent, and an analysis of all of the hyperparameters involved, see
Yoshua Bengio's tutorial ``Practical recommendations for gradient-based
training of deep architectures." (Ref.~\citenum{bengio2012practical}.)

\subsection{Autoencoders \& Deep Neural Networks}

A deep neural network is a neural network with more layers than is
traditionally used. Layered neural networks are also known as multi-layer
neural networks or multi-layer perceptrons (MLPs), although the latter is
a misnomer. A single layer MLP can be formally described as a function $f
: \mathbb{R}^D \to \mathbb{R}^P$ parameterized by $\theta = (\mathbf{W}_0,
\mathbf{b}_0, \mathbf{W}_1, \textbf{b}_1, \sigma_0, \sigma_1)$. Where $D$ is
the number of dimensions of the input $\mathbf{x} \in \mathbb{R}^D$ and $P$ is
the number of dimensions of the output layer.

$$
f(\mathbf{x}) = \sigma_1(\mathbf{W}_1(\sigma_0(\mathbf{W}_0\mathbf{x} + \mathbf{b}_0)) + \mathbf{b}_1)
$$

The function $\sigma : \mathbb{R} \to \mathbb{R}$, is referred to as
the \textit{activation function}, or, \textit{nonlinearity}, because it often
is a non-linear function such as tanh, sigmoid, or a rectified linear unit
(ReLU)\footnote{Rectified Linear Units are defined as $r(x) = max(0, x)$}. The
parameter set, $\theta$ is then optimized to minimize the training loss
$\mathcal{L}$ using SGD, L-BFGS, or another optimization
algorithm.\cite{byrd1995limited}

\subsubsection{Unsupervised Learning}

Unsupervised feature learning is a set of unsupervised learning techniques that
transform the input into a form that is easier to work with for other tasks such as clustering, classification, or
regression.\cite{bengio2013representation} Autoencoders are one form of
unsupervised learning algorithm.

Formally, an autoencoder is a set of two functions: an \textit{encoder} and
a \textit{decoder}. The encoder function parameterized by $\theta$, $f_\theta$,
maps the input vector $\mathbf{x} \in \mathbb{R}^{D}$ onto a hidden layer, or
\textit{code}, $\mathbf{y} \in \mathbb{R}^H$ and the decoder function
parameterized by $\theta'$, $g_{\theta'}$, maps the code $\mathbf{y}$ onto
a reconstructed vector $\mathbf{z} \in \mathbb{R}^{D}$.\cite{bengio2009learning} The job of the learning
algorithm is to modify the parameters $(\theta, \theta')$ of the encoder and
decoder to better reconstruct the original input from the code.  Modifications to the
autoencoder scheme include denoising autoencoders, contractive autoencoders,
and sparse autoencoders.\cite{vincent2008extracting, vincent2010stacked, rifai2011contractive, le2012} Denoising autoencoders are a simple modification of
a standard autoencoder. They add noise to the input vector $\mathbf{x}$ to get
$\mathbf{\widetilde{x}}$ and then attempt to reconstruct the original vector
with the noisy input $\mathbf{z}
= g_{\theta'}(f_{\theta}(\mathbf{\widetilde{x}}))$.\cite{vincent2008extracting}

The loss functional for an autoencoder $\mathcal{L}$ can range from squared
error $\mathcal{L}(\mathbf{x}, \mathbf{z}) = ||\mathbf{x} - \mathbf{z}||^2$ to
the cross-entropy of the reconstruction $\mathcal{L}(\mathbf{x}, \mathbf{z}) = \sum_{i=1}^d[ \mathbf{x}_i \log \mathbf{z}_i + (1-\mathbf{x}_i) \log(1-\mathbf{z}_i)]$. The choice of the loss functional depends on the assumed
distribution of the input code, $\mathbf{y}$.\cite{bengio2009learning}
Autoencoders can be used to \textit{pretrain} (initialize the weights of) a DNN
which is then \textit{fine-tuned} with labeled data, i.e. trained in
a supervised manner.\cite{vincent2008extracting, vincent2010stacked}  This
technique is known as unsupervised pretraining.\cite{bengio2007greedy}

Creating your own unsupervised face dataset is relatively
straightforward due to the robustness of face detection software. All that is
needed to form an unlabeled face dataset is a large collection of photos of
people and a face detector. With the torrent of images available online, it's
easy to gather a large unlabeled dataset for pretraining DNNs.

Because input images to autoencoders and DNNs are flattened, mapped from
$\mathbb{R}^{w \times h} \to \mathbb{R}^{wh}$, autoencoders and
multi-layer NNs lose out on the inherent local 2D structure of images.
Convolutional neural networks (CNNs), on the other hand, take advantage of this
2D structure.

\subsection{Convolutional Neural Networks}

Convolutional neural networks date back to 1980 with Fukushima's
Neocognitron.\cite{fukushima1980neocognitron} They were improved and
successfully applied to handwritten digit recognition by Yann LeCun in the late
90s; early applications to face recognition appeared around the same time.
\cite{lawrence1997face}

Unlike DNNs, which operate by performing a dot product between 1D input vectors
and the network's weight matrix followed by an element-wise nonlinearity,
$\sigma$, $\sigma(\mathbf{W}\mathbf{x} + \mathbf{b})$, a CNN operates by
performing a 2D convolution of the filters in its filter bank with a 2D input
vector: $\sigma(\textbf{W} \ast \textbf{x} + \textbf{b})$. Nearly every method
that performs well on LFW utilizes CNNs. (See Table 1.) Another major advantage
of CNNs falls out from the nature of the convolution operation: a linear
translation in the input data causes a linear translation in the feature map.
This provides some degree of translation-invariance which is not found in DNNs
and autoencoders.

\section{The State-of-the-art in Facial Recognition}

The standard pipeline for facial recognition has changed drastically over the
past few years. We've seen a transition from hand engineered features to
learned features, a transition from face-specific alignment to rough alignment
and centering, and a transition from datasets with tens of thousands of images
to datasets with hundreds of millions of images. The various phases of this
transition are shown below:

\begin{enumerate}
\item No alignment needed, hand engineered features, dataset size $\approx$ 1e3, dataset gathered in highly controlled lab environment.\cite{ahonen2006, samaria1994parameterisation, belhumeur1997eigenfaces}
\item Domain specific alignment, hand engineered features, SVMs, dataset size $\approx$ 1e4.
\item Domain specific alignment (face-specific frontalization and deep funneling), learned features, dataset size $\approx$ 1e6.\cite{taigman2013, huang2012learning}
\item Domain specific alignment (rough alignment), learned features, dataset size $\approx$ 1e7.\cite{schroff2015facenet}
\end{enumerate}



\begin{table}[h]
\label{tab:methods}
\caption{State-of-the-art methods on LFW at time of publishing. (Sorted by accuracy descending.)}
\begin{center}
\begin{tabular}{|l|l|l|l|l|}
\hline \textbf{Name} & \textbf{Method} & \textbf{Images (Millions)} & \textbf{Accuracy} \\
\hline
Baidu\cite{andrewng2015}\textsuperscript{(Announced)} & CNN & - & 0.9985 $\pm$ -\\
\hline
\textbf{Google FaceNet\cite{schroff2015facenet}} & \textbf{CNN} & \textbf{200.0} & \textbf{0.9963} $\pm$ \textbf{0.0009}\\
\hline
DeepID3\cite{sun2015deepid3} & CNN & 0.29 & $0.9953 \pm 0.0010$ \\
\hline
MFRS\cite{faceplusplus2015} & CNN & 5.0 & $0.9950 \pm 0.0036$ \\
\hline
DeepID2+\cite{deepid2plusyisun2014} & CNN & 0.29 & $0.9947 \pm 0.0012$ \\
\hline
DeepID2\cite{sun2014deep} & CNN & 0.16 & $0.9915 \pm 0.0013$ \\
\hline
DeepID\cite{deepidsun2014} & CNN & 0.2 & $0.9745 \pm 0.0026$ \\
\hline
DeepFace\cite{taigman2013} & CNN &  4.4 & $0.9735 \pm 0.0025$ \\
\hline
FR+FCN\cite{frfcn} & CNN & 0.087 & $0.9645 \pm 0.0025$ \\
\hline TL Joint Bayesian \cite{cao2013practical} & Joint Bayesian & 0.099 & $0.9633 \pm 0.0108$ \\
\hline
High-dim LBP \cite{chen2012bayesian} & LBP & 0.099 & $0.9517 \pm 0.0113$ \\
\hline
\end{tabular} \end{center} \end{table}

Despite the move from engineered features to deep CNNs, all of the
state-of-the-art methods still utilize face specific alignment techniques. This
ranges from rough centering in Schroff, Florian and Kalenichenko 2015 to the
use of a 3D face mask estimate and re-projection of the 2D image as in Hassner,
Tal, et al.~2014.\cite{schroff2015facenet, hassner2014effective}

\section{DATASETS} \subsection{Datasets today}

\begin{table}[h] \caption{An overview of known public and private face
datasets. (Sorted by images descending.)} \label{tab:datasets} \begin{center}
\begin{tabular}{|l|l|l|l|} \hline \textbf{Dataset} & \textbf{Identities} &
\textbf{Images (Millions)} & \textbf{Availability} \\ \hline Google Face
Dataset\cite{schroff2015facenet} & 8,000,000 & 260+ & \textbf{Private} \\
\hline Megavii Face Classification (MFC) \cite{faceplusplus2015} & 20,000 & 5.0 &
\textbf{Private} \\ \hline Social Face Classification (SFC) \cite{taigman2013}
& 4,030 & 4.4 & \textbf{Private} \\ \hline CASIA-WebFace \cite{casiayi2014} &
10,575 & 0.494 & Public \\ \hline CelebFaces \cite{celebfacesun2014} & 10,177 &
0.202 & \textbf{Private} \\ \hline CACD \cite{cacdchen14cross} & 2,000 & 0.163
& Public \\ \hline WDRef \cite{chen2012bayesian} & 2,995 & 0.099 & Public
(features only) \\ \hline LFW \cite{lfw} & 5,749 & 0.013 & Public \\ \hline
\end{tabular} \end{center} \end{table}

Nearly all of the top algorithms on the LFW benchmark are trained using outside
data. LFW offers just 13,233 images. That amount of data is insufficient to
properly train modern deep networks.  Table 1 shows the top
results of LFW, the accuracy achieved for the face verification task, and the
amount of data used to train the system.

Table 1 shows not only the dominance of deep convolutional neural networks but
also the importance of large datasets for training these networks. Table 2
shows the private nature of the largest face datasets. This trend is
disconcerting. If these large datasets remain private, it's possible that
progress in facial recognition research might be restricted to those with
access to large amounts of proprietary data. According to estimates in 2013 by
Kleiner, Perkins, Caufield \& Byers (KPCB), over 1.8 billion images are
uploaded and shared per day on mobile photo sharing
networks.\cite{meeker2013}\footnote{KPCB's analysis only covered companies in
the mobile photo sharing market. Other companies, such as Tencent (WeChat),
Google, Dropbox, Apple, and Yahoo also upload millions of photos on a daily basis.
However, those numbers remain unpublished.}

\begin{table}[h] \caption{Billions of images are uploaded and shared every
day.} \label{tab:mobile} \begin{center} \begin{tabular}{|l|l|} \hline
\textbf{Company} & \textbf{Daily Uploads} \\ \hline Snapchat & 700 M / day \\
\hline WhatsApp (Acquired by Facebook in 2014) & 500 M / day \\ \hline Facebook
& 350 M / day \\ \hline Instagram (Acquired by Facebook in 2012) & 60 M / day
\\ \hline \end{tabular} \end{center} \end{table}

\begin{figure}[h]
   \begin{center}
   \begin{tabular}{c}
   \includegraphics[height=200px]{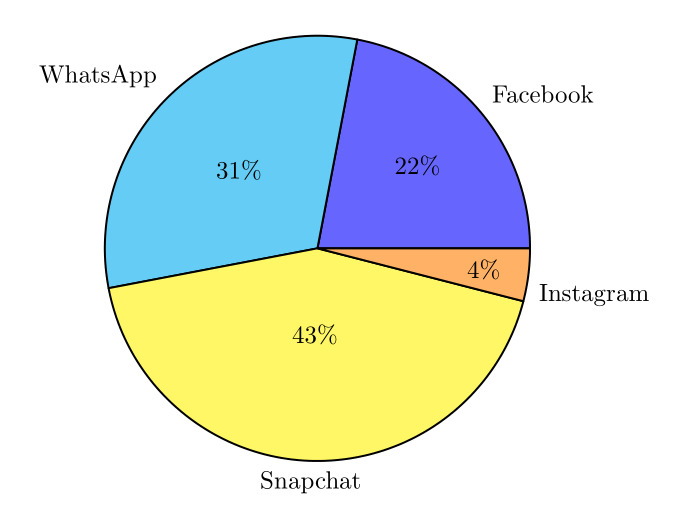}
   \end{tabular}
   \end{center}
   \caption[Images uploaded per day]{Who's uploading the most?}
\end{figure}

Publicly available datasets with tens of millions of images could be compiled
and would remove a significant roadblock for progress in the field.  The
optimal size and structure of such a dataset is outside the scope of this
paper, however, Schroff, Florian, et~al.~(See
Ref.~\citenum{schroff2015facenet}) suggest that the current generation of deep
networks benefit from training sets with tens of millions of images and
saturate thereafter.\cite{schroff2015facenet} The results of their study on the
effect of training data on performance is reproduced below in Table 4.

\begin{table}[h]

\caption{The effect of training set size on the performance of a model after
700 hours of training on 96x96 pixel input images.\cite{schroff2015facenet}}

\label{tab:google}
\begin{center}
\begin{tabular}{|l|l|}
\hline
\#images & validation rate \\
\hline
\hline
2.6M & 76.3\% \\
26M & 85.1\% \\
52M & 85.1\% \\
260M & 86.2\% \\
\hline
\end{tabular}
\end{center}
\end{table}

\subsection{The problem with today's datasets and benchmarks}

\subsubsection{False Accept Rates (FARs) that are too high.}

Yi et al.~2014 noted that the accuracy of state-of-the-art methods may be
saturating LFW; the author shares this opinion.\cite{casiayi2014}  Yi et
al.~suggest BLUFR, which has a focus on lower false accept rates, as a
more challenging alternative to LFW. While a benchmark which focuses on lower
FARs is needed, a look at the current state-of-the-art also shows the need for
a very large scale public dataset.

\subsubsection{Lack of variety and poor generalization.}

Previous generation datasets like AT\&T and Yale were captured under highly
controlled laboratory environments.\cite{samaria1994parameterisation,
belhumeur1997eigenfaces} Datasets such as LFW and CACD claim to be ``in the
wild", but are taken almost exclusively by professional photographers with
DSLRs in well lit environments. Training on such datasets will likely lead to
poor generalization error when the models are confronted with a less
constrained operating environment such as a photo stream from a mobile phone.

\subsubsection{Not enough data.}

Deep neural networks require large amounts of data, preferably tens of millions
of images.\cite{schroff2015facenet} As demonstrated in Table 1, all top
performing methods on LFW take advantage of large, supplementary datasets. LFW
by itself is simply not enough data given the capacity of modern deep
architectures.

\subsection{Proposal for a new kind of facial recognition dataset}

Much like the transition that occurred from Caltech 101 to ImageNet within the
visual object recognition field, the author posits that a similar transition
will occur within the field of face recognition from LFW to a large scale
dataset and corresponding challenge. The companies featured in
Table~\ref{tab:mobile} would be in a solid position to create and publish such
a dataset.

\section{CONCLUSIONS}

Deep learning has already been integrated into most state-of-the-art facial
recognition pipelines. This shift has lead to a massive increase in the accuracy
of facial recognition systems and has caused the current ``standard" benchmark
for face recognition, LFW, to become saturated.  In addition, the data
requirements for deep networks highlights the need for a new, very large scale (tens
of millions of images), public dataset for face recognition research.

\acknowledgments

This work was supported by Lambda Labs. (\url{https://lambdalabs.com})


\bibliography{bibliography}   
\bibliographystyle{spiebib}   

\end{document}